\documentclass[final, twocolumn]{elsarticle}
\usepackage{amssymb}
\usepackage{url}
\usepackage{color}

\journal{JAG}
\begin{document}
\begin{frontmatter}

\title{Example-Based Explainable AI and its Application for Remote Sensing Image Classification}

\author[rikkyo,mamezousdb]{Shin-nosuke Ishikawa}
\author[mamezousdb]{Masato Todo}
\author[rikkyo]{Masato Taki}
\author[rikkyo]{Yasunobu Uchiyama}
\author[mamezousdb]{Kazunari Matsunaga}
\author[mamezousdb]{Peihsuan Lin}
\author[mamezousdb]{Taiki Ogihara}
\author[mamezou]{Masao Yasui}

\address[rikkyo]{Graduate School of Artificial Intelligence and Science, Rikkyo University, Tokyo 171-8501, Japan}
\address[mamezousdb]{Strategic Digital Business Unit, Mamezou Co., Ltd., Tokyo 163-0434, Japan}
\address[mamezou]{Mamezou Co., Ltd., Tokyo 163-0434, Japan}

\begin{abstract}
We present a method of explainable artificial intelligence (XAI),
``What I Know (WIK)'',
to provide additional information to verify the reliability of a deep learning model by showing an example of an instance in a training dataset that is similar to the input data to be inferred and demonstrate it in a remote sensing image classification task. One of the expected roles of XAI methods is verifying whether inferences of a trained machine learning model are valid for an application, and it is an important factor that what datasets are used for training the model as well as the model architecture. Our data-centric approach can help determine whether the training dataset is sufficient for each inference by checking the selected example data. 
If the selected example looks similar to the input data, we can confirm that the model was not trained on a dataset with a feature distribution far from the feature of the input data. 
With this method, the criteria for selecting an example are not merely data similarity with the input data but also data similarity in the context of the model task. Using a remote sensing image dataset from the Sentinel-2 satellite, the concept was successfully demonstrated with reasonably selected examples. This method can be applied to various machine-learning tasks, including classification and regression. 
\end{abstract}


\end{frontmatter}


\section{Introduction}
Machine learning techniques, including deep learning, often considered a subset of artificial intelligence (AI) technologies, are widely used in many research and business fields nowadays, especially in combinations with so-called ``big data,'' including satellite remote sensing data. Deep learning methods have rich expressive power comparable to or even superior to human cognitive abilities in many tasks\cite{lecun2015}.
Several applications of deep learning techniques using satellite data have been proposed, including astronomy, planetary sciences, and earth observations, and research efficiency and quality improvements are expected
\cite{ishikawa2021, cheng2020, zhang2016, kussul2017}.

In contrast, the inference process of deep learning models is often regarded as a black box, owing to the vast number of parameters and computational complexities. 
This might obstruct the decision of production implementations of AI systems. Explainable artificial intelligence (XAI) is a field in AI research that explains and interprets machine-learning models to avoid this situation\cite{barredoarrieta2020, molnar2022}. 
Discussions have also begun on application of XAI techniques to Earth observation \cite{gevaert2022, kakogeourgiou2021}.
When applying XAI methods to real-world tasks, we must be careful to clarify the purpose of the explanation.

Local Interpretable Model Explanations (LIME)\cite{tulioribeiro2016} and SHapley Additive exPlanations (SHAP)\cite{lundberg2017} are representative XAI methods for importance analyses of explanatory variables. 
LIME explains models using a local linear approximation and outputs values indicating the contribution of each explanatory variable to a given inference. 
SHAP also outputs the importance of each explanatory variable for the inference using Shapley values based on game theory.
For image data, Gradient-weighted Class Activation Mapping (Grad-CAM)\cite{selvaraju2016} is a widely used XAI method that shows areas of importance in an input image for each inference of a convolutional neural network (CNN) model by calculating the gradient of the last convolutional layer in the model. This method can also be categorized as an importance analysis XAI method. 
These methods are valuable both for explaining the behavior of deep learning models to stakeholders and improving performance, while results from different methods are not always consistent.

In tasks where feedback from information obtained by the XAI is valuable, importance analysis methods for explanatory variables are useful, such as in cases of risk management\cite{wen2021} and quality control in industries\cite{senonar2021}. 
Using the importance analysis methods, we can investigate which input data are significant for each inference and whether those inputs affect the results positively or negatively. 
Following these explanations, we can control the input variables to avoid risks or improve the quality. 
Additionally, risk or quality control of image data is possible by identifying and removing spots related to undesirable conditions using methods such as Grad-CAM.

Despite their advantages, we should note that these XAI methods only provide explanations or interpretations of inferences or models, not evidence or rationale for inferences.
Because the XAI explanation does not show the inference process, it is incorrect to conclude that the important explanatory variables are the reason why the model is inferred. Variable importance is calculated using approximations (e.g., LIME) or counterfactual assumptions (e.g., SHAP). Deep learning and other state-of-the-art models can reproduce complicated nonlinear relations, and they should not be undermined by the trivialization of the models. In this context, we point out several risks of utilizing XAI methods and suggest using interpretation-friendly models instead\cite{rudin2018}. We must understand what is actually calculated and what is not obtained from XAI methods. 

The motivation for using XAI methods is not only for feedback to control the input data, but also to verify whether the inference results are reliable. To meet this demand, importance analyses of explanatory variables are insufficient because knowing which explanatory variables are important does not clarify how the model determines those variables to be important. In other words, we may have questions, such as \textit{why predict this when this variable is high.} In thfis view, one of the possible approaches to answer such questions is to show the data trained by the model. Compared with simplifying the model, it is more accurate to explain that the model is trained using this dataset and find an abstract pattern for inferences. This approach is related to a ``data-centric'' perspective of machine learning applications. The training dataset is an important factor in machine learning model performance along with model architectures and training algorithms, and the data-centric approach focuses on the dataset, unlike the conventional model-centric approach. 
It would be valuable to show a summary of the training dataset, such as the number of instances and statistics for the explanatory and objective variables. When and how the data are collected could also be important reference information to verify whether we can assume that the training and input data to be inferred in the production release are from the same population.

Showing an example of instances from a training dataset would be a beneficial data-centric explanation of the model inferences \cite{molnar2022}. In particular, an example similar to the inferred target data would be a valuable reference to verify whether the model is trained on a dataset sufficiently close to the target data\cite{hanawa2021}. If the most similar example is too far away from the target, we can interpret that it is invalid to apply the model to the target data, and the inference result is suspicious.

In this study, we present a data-centric XAI method that shows an instance in the training dataset similar to the target data to be inferred by a deep learning model. We demonstrate our method using a remote-sensing image classification task and discuss effective cases.

\section{Methodology}
We present a method
named What I Know (WIK)
 for evaluating similarities in the data instances using the model outputs and calculated values of the last layer before the output layer in a deep-learning model. Although there are several metrics to quantify data similarities, such as norms and cosine similarity, our approach is not to measure similarities of input data themselves but to evaluate similarity after data abstraction by the model. 
Therefore, WIK's results are model-dependent, and it is possible that the most similar data instances are not the same when we evaluate similarities with the different models. This corresponds to a similarity evaluation in the context of a task for which the model is trained. We will show the difference between the traditional similarity measurement and our context-based method by providing examples in the discussion section. 

\begin{figure*}[!t]
\centering
\includegraphics[width=5.4in]{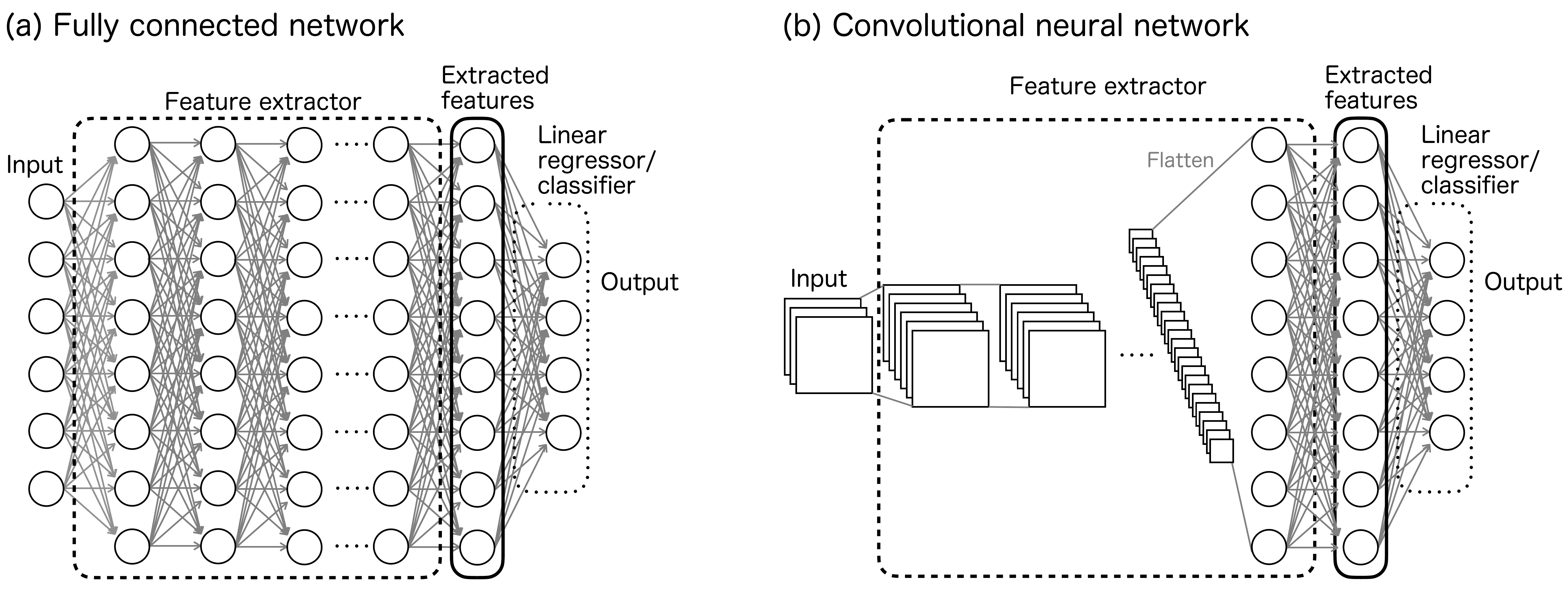}
\caption{Concept of the interpretation that the neural network consists of the feature extractor and a simple linear classifier/regressor for the extracted features, 
for (a) a fully connected network and (b) a CNN.
}
\label{fig_concept}
\end{figure*}

In this method, the similarity is measured by the $L^2$-norm between the vectors of the last layer before the output layer as
\begin{equation}
\sqrt{\sum _{i=1} ^{n} \left( f_i - g_i \right)^2}
\end{equation}
where $f_i$ and $g_i$ are the i-th components of the vectors and $n$ is the dimensionality of the vectors, 
 by considering the part of the model before the last layer as a feature extractor. 
 Figure~\ref{fig_concept} shows the concept of an idea that a deep learning model consists of a feature extractor and a simple linear classifier or regressor. In the layer before the output layer, features of the input instance for the task the model trained for are extracted, and we can compare input data instances using those features. 
As shown in Fig.~\ref{fig_concept}, this concept is applicable to CNN models, in addition to fully connected networks.
 The idea of using the output of the layer before the last layer is used for some applications, such as the Fr\'echet Inception Distance \cite{heusel2017} for measuring the similarity of data distributions mainly for evaluating generative networks. We investigated several similarity metrics in a multidimensional space, such as cosine similarity, and concluded that the $L^2$-norm provides the most reasonable results. In addition, the $L^2$-norm evaluation achieved the best result in the feature similarity measurements compared by Hanawa et al.\cite{hanawa2021}. 
We also provide results with other similarity metrics in the next section. 

\begin{figure*}[!t]
\centering
\includegraphics[width=5.4in]{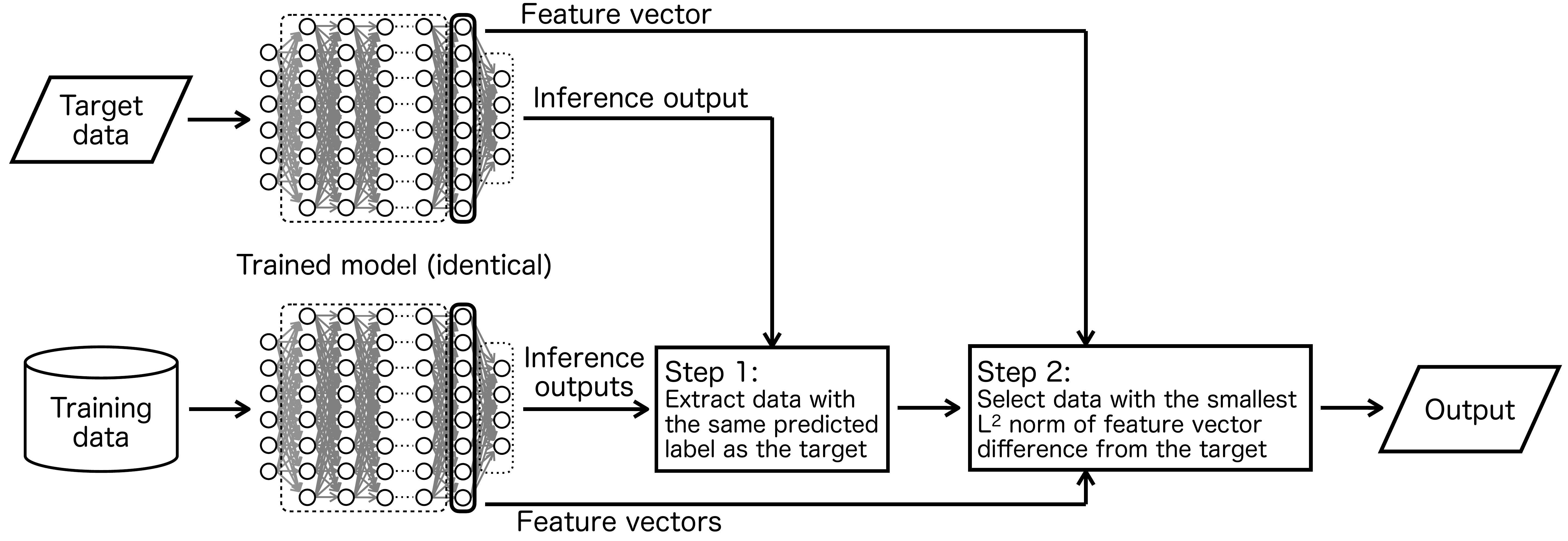}
\caption{
Flow of selecting the data instance most similar to the target data among the training data.
}
\label{fig_flow}
\end{figure*}

The similarity evaluation with the last layer before the output layer does not use the weights of the connections to the output layers, and we applied a two-step similarity evaluation 
as shown in Fig.~\ref{fig_flow}. 
These weights provide important information for deriving inference results and are necessary for understanding how the model interprets the extracted features. The first step of the similarity evaluation is verifying the inference results, which the model concludes. Here, we focus on classification tasks and set the criterion that similar instances should be inferred as an identical class. In our future work, we will present a method for regression tasks. In the second step, we measure the $L^2$-norm of the vectors from the last layer before the output layer. A method of similarity measurement with the $L^2$-norm of the layer before the output has already been shown in a previous study\cite{hanawa2021}. In that work, it was pointed out that data instances that are evaluated as similar are not always classified as the same class. While it is not evident that we must avoid the mismatch for similarity evaluation XAI methods, such a situation does not occur in our method, owing to step 1. 

A similarity value calculated by WIK can also be output to evaluate the distances between instances in the feature space.
It would be useful to compare similarity values for different instances in the test dataset to understand the difference between the training and test datasets.
The similarity values are also useful to investigate how a model update affect the inference of the target instances.
Note that the similarity values of models trained for different tasks should not be compared because the scales of the similarity values are model-dependent.

To demonstrate the WIK method shown above, we use an example of remote sensing image classification using the EuroSAT dataset with the images taken by the Sentinel-2 satellite\cite{helber2017}. EuroSAT is a dataset with multi-spectral images in a 64$\times$64 pixel format and is labeled as ten classes of scenes, such as forest, river, industrial, and residential. EuroSAT has two types of datasets with difference sets of wavelength bands. Here, we use the EuroSAT visible light observation dataset with red, green, and blue color channels. We constructed a standard LeNet-type CNN with three sets of two convolutional layers and a pooling layer in sequence, followed by fully connected layers. The output layer had ten nodes corresponding to 10 classes, and the last layer before the output layer had 32 nodes. 

We split the visible light EuroSAT dataset into 80\% and 20\% for training and testing, respectively, and trained the model with the training dataset. 
The data were randomly divided between the training and testing sets, and both were confirmed to be class-balanced.
We evaluated the model using the test dataset and obtained an accuracy of 95.7\%. While this performance is not as good as the state-of-the-art results shown in a recent study\cite{gomez2021}, the purpose of the demonstration in this article is to show an example of the usage of our XAI method for real-world applications. This is a realistic example of a model that can be constructed in finite time.
While it is true that models trained by fine-tuning or transfer learning from a model pre-trained by a large dataset are often used in practical applications, we selected a simpler model structure to highlight the XAI method presented here.

\section{Results}
\begin{figure*}[!t]
\centering
\includegraphics[width=5.4in]{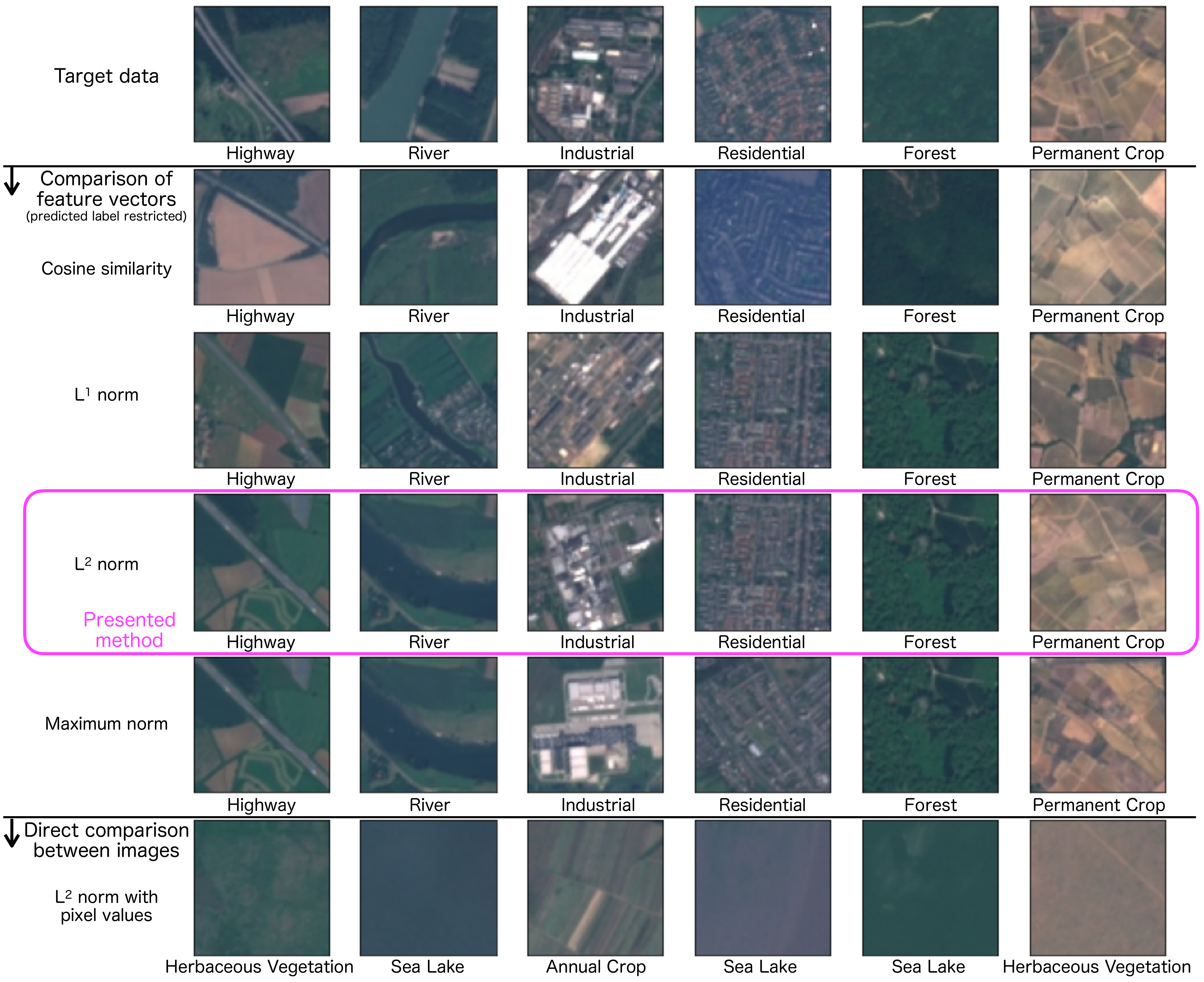}
\caption{
Selected images by the presented XAI method with different similarity measurements (second to fifth rows) for the input images (first row) and images with the smallest $L^2$-norm of the pixel values (sixth row). 
A label is shown for each image. All the predicted labels in the images shown here are correct. In the proposed method, the label of the selected image is necessarily the same as that of the original image.
}
\label{fig_result1}
\end{figure*}

We applied the example-based XAI method WIK presented in the previous section to the trained model. Examples of the input data instances to be inferred from the test dataset and those similar to the input data selected from the training dataset by the method are shown in Fig.~\ref{fig_result1}. The six images in the first row show the input data from the test dataset, 
and the images from the second row to the fifth row are images from the training dataset that the model considered being similar with different similarity metrics, including cosine similarity
\begin{equation}
\frac{\vec{f} \cdot \vec{g}}{\|\vec{f}\| \| \vec{g}\|},
\end{equation}
$L^1$-norm
\begin{equation}
\sum _{i=1} ^{n} | f_i - g_i |,
\end{equation}
 $L^2$-norm, and maximum norm
 \begin{equation}
\max _{i} | f_i - g_i |.
\end{equation}
Apparently, the corresponding image pairs of the target data and the $L^2$-norm result (fourth row) 
look similar to structures of similar spatial scales, colors, and textures, although the positional relationship is not exactly the same. The highway images show almost straight highways with similar widths surrounded by rural scenery, and the rivers in the river images are similar. The industrial images have several similar buildings with similar sizes, shapes, and colors, and the same relationship can be seen in the residential images. While the contrast of the forest images is different to some extent, the forest and permanent crop image pairs have similar spatial scales and colors. This result indicates that abstracted features, similar to text describing an image, are successfully extracted by the extractor part of the model, although it is difficult to evaluate such similarities using a predetermined mathematical formula. Owing to step 1 of our method, the predicted labels for the selected images are the same as those of the original images. Among the same-label instances in the training dataset, the selected images have the most similar features.

A comparison of the results with the different similarity metrics shown in Fig.~\ref{fig_result1}, particularly for the river and industrial images, reveals that t$L^2$-norm provides the most reasonable result.  
The widths and curvatures of the river images are similar to those of the target image for the $L^2$- and maximum norm results. 
The sizes, shapes, and colors of the buildings in the industrial image with the $L^2$-norm are similar to those in the target data compared to the results with the other metrics. 
While 3 out of 6 images are the same for the $L^2$-norm result and maximum norm, the spatial scales of the structures in the residential and industrial images are more similar to the target data by the $L^2$-norm.
Based on these results, we use the $L^2$-norm as the best similarity measurement method for WIK in the following discussion.

With the result from WIK, we can investigate what data were used for training the model and what data instance pairs the model captures as similar. The purpose of the XAI methods is to explain model inferences to humans, and we should consider how we receive the results. If we consider the presented pair to be reasonably similar, it shows that the model extracts features without being counterintuitive. In other words, we can verify whether abstraction by the model is reasonable and consistent with that of humans. In addition, we can verify whether the model was trained on the dataset sufficiently covering the feature space to estimate the positioning of the target. If we do not feel the pair is similar, we should suspect that the training dataset is insufficient in the feature space around the target or that the model is insufficiently trained. In general, the inference result is unreliable in such situations. 

\begin{figure}[!t]
\centering
\includegraphics[width=3.1in]{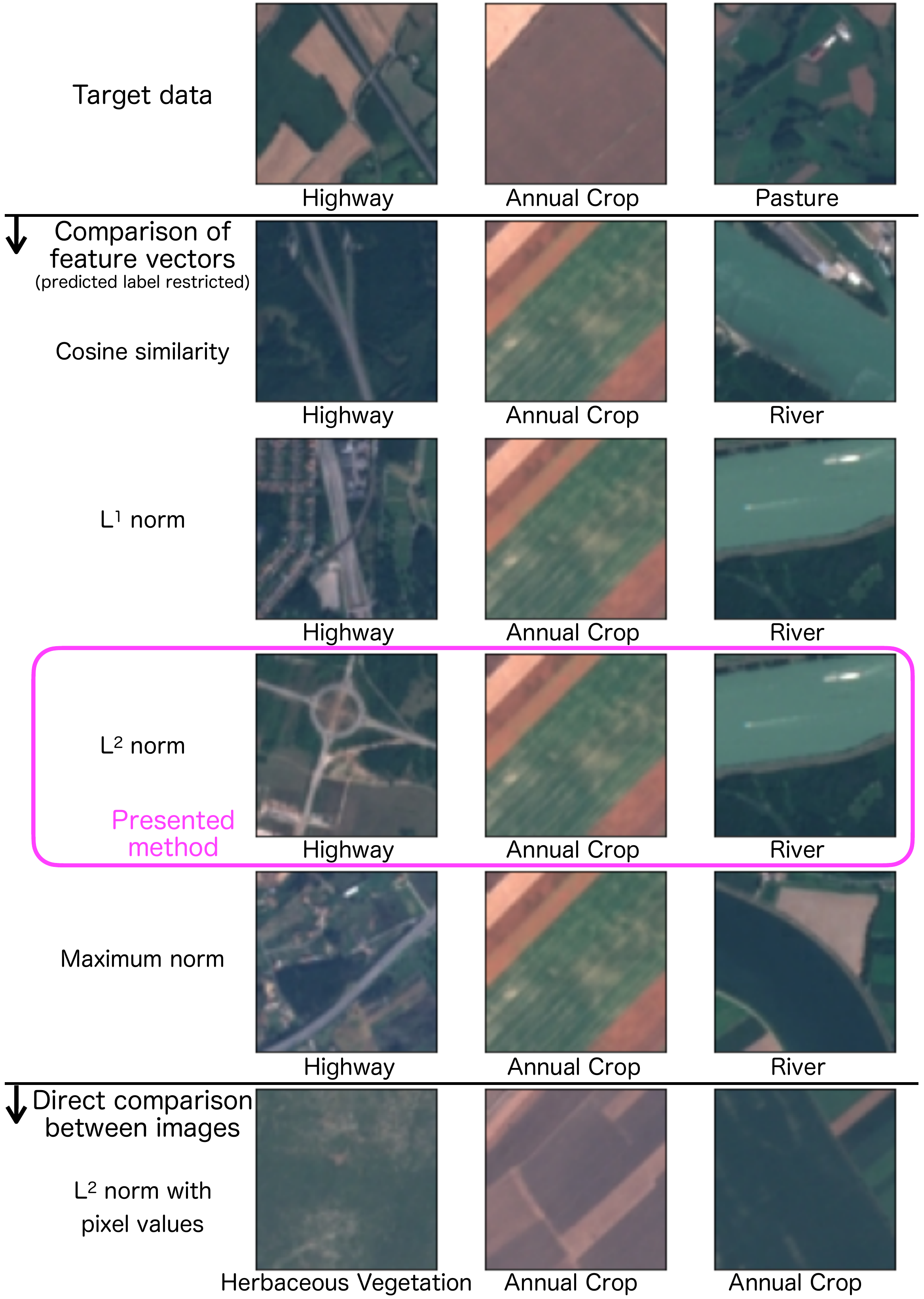}
\caption{
Selected images by the presented XAI method with different similarity measurements shown as in Fig.~\ref{fig_result1} for cases in which the images appear dissimilar. 
A predicted label is shown for each image. The predictions of the images labelled as showing a river are incorrect; these images should have been correctly labeled as pasture.
}
\label{fig_result2}
\end{figure}

Figure~\ref{fig_result2} shows results similar to those in Fig.~\ref{fig_result1}, but for the cases in which the images do not look so similar. 
As we can see, the images are not similar for other metrics as well as for the $L^2$-norm results.
These are rare cases with the model and the training dataset shown here.  If these results are dominant, we would suspect that the quality of the model is not good.
In the highway images, a junction is observed on the result in the fourth row.  The model might interpret that the highway target data shows a crossroad and it is similar to the highway junction. 
The selected result for the annual crop image has a different color from the target, and we can see that colors are not important to the model for areas of this type. 
For the target data showing a pasture, the inference result is incorrect and the data is classified as a river. This causes results to be selected from the incorrect class, thus leading to inconsistency between the target data and the results. 

We are unaware of any effective way to evaluate example-based XAI methods because the value of those methods depends on whether the users satisfy the outputs. While one of the approaches to verify the self-consistency of the methods is shown\cite{hanawa2021}, that metric inevitably becomes 100\% with our method because it is requested in step 1 of the process. For real-world applications, it is critical that stakeholders understand the method's positioning and are persuaded to use it.

\section{Discussion}
We successfully showed that the presented XAI method WIK could provide information to verify that the AI model is appropriately trained with the data-centric approach, and this result cannot be achieved by simply using mathematical distance metrics for input data. The images in the sixth row in Fig.~\ref{fig_result1} are also images from the training dataset but are selected by the criterion that the $L^2$-norm 
of the pixel values of the images themselves, 
not of the features extracted by the model used in the WIK method, 
with respect to the target data is the smallest. 
Although it is also a similarity metric, the selected images are different from those selected by the presented XAI method shown in the fourth row. The spatial structures and contrast are weaker than the original images and the image in the second row, and the images are relatively smooth and uniform. This is caused by the manner in which the $L^2$-norm is calculated. 
For spatially structured features, the $L^2$-norm 
of the pixel values rather than that of the feature vectors
is high if the pattern does not match perfectly. 
It is unrealistic that there is an image with a perfectly matching structure pattern, and the $L^2$-norm cannot extract abstract features of images related to spatial structures. Therefore, the $L^2$-norm measurement of the original images is unsuitable for the XAI method. By comparing the images in the second and third rows, it is clearly shown that the XAI method described in the previous section can successfully select images with similar features in the context of the task to classify what is in the image. 

From the results of WIK, as shown in Fig.~\ref{fig_result1}, we can check whether each inference is suspicious during the production phase of AI model applications. When the inference results are used for making decisions, we can use the XAI results as discussion material in addition to the inference results themselves. Even when a machine learning model is used as part of a system for automation, although there might not be a natural way to include the XAI method in the system itself, the XAI result is available for performance monitoring. This perspective is closely related to the idea of MLOps for discussions on how to maintain machine-learning models. How we can detect degradations of machine learning models is under discussion, and there is a possibility that example-based XAI methods, such as those presented here, can help investigate the relationship between the distribution of recent target data and the model. 
As the performance of the model degrades, instances selected by the WIK method might become less intuitively similar owing to discrepancies between the latest input data and the training dataset acquired in the past. Similarity values may also worsen with degradation.

Because the purpose of WIK demonstrated here is to obtain additional information to verify the reliability of the model inferences, this method is effective in the production and proof-of-concept phase. We may have to suspect 
that related instances that should not be separated, or even exactly the same instance, are included in both of the training and validation datasets
if the pair is \textit{too} similar. 
Conversely, if the pair does not look similar, we could improve the generalization performance of the model by investigating its origin. There are several possible explanations for this finding. The variation of the training dataset might be insufficient, or hyperparameters might not be set to accurate values. In addition, there is a possibility that this was caused by only a simple mistake. The proposed XAI method can help avoid such inappropriate situations. 

If a large number of instances, such as the entire test dataset, need to be evaluated with WIK, the result can be summarized by performing a visual inspection of the selected instances with the highest and lowest similarity values. Identical instances can be identified in the training and test datasets owing to their complete similarity. The lowest similarity result indicates an area with few training data instances in the feature space. Investigating the highest and lowest similarity instances for each class can help understand the distribution of the training data in the feature space.

As we have shown in the introduction, the WIK method presented in this article is not necessarily effective for all purposes of XAIs. In real-world applications, selecting an appropriate method for each case is important. If the analysis focuses on which explanatory variable is important, importance analysis methods for the input data, such as LIME or SHAP, would be an appropriate choice. In such a case, WIK is ineffective. Conversely, if the reliability of the AI model is a concern, the presented method provides a unique perspective for model validity investigation using a data-centric approach. 
The importance analysis methods provide a component analytical approach, whereas WIK provides reference data the model trained on. 
Therefore, the importance analysis methods and WIK are complementary in understanding machine learning models from multiple perspectives.

Decision-making based on AI predictions could be a major application of the XAI method. We must be careful about risks when making decisions, and it is dangerous to use only inference results in many cases. In addition to the fact that AI inferences are not always perfect, we should be careful if validations with performance metrics do not guarantee prediction accuracy through false implicit assumptions. This could be caused by the difference in the data distributions between the training dataset and the new data to be inferred. Few XAI methods effective for this situation have been suggested, and the method presented here is a candidate to overcome this difficulty. 

\section{Conclusion}
We presented an XAI method WIK to investigate the reliability of deep learning models by showing a data instance that has the most similar features in the context of the task of the model.
Our approach differs from the XAI methods with the importance analysis of explanatory variables and helps to intuitively understand how the model infers. 
This method can verify the training data, an important component of machine learning, along with the models, from a data-centric perspective. 
Experimental results on a remote sensing image classification task with the EuroSAT dataset, we confirmed that the WIK method performs well by selecting reasonably similar instances compared to the input data. 
With this method, applications such as AI reliability monitoring, validity checks for AI developments, and more careful decision-making are expected.

\section*{Acknowledgments}
We would like to thank Prof. Katsuki Fujisawa and graduate students in his laboratory to have a discussion and give us useful comments. 

\bibliographystyle{elsarticle-num} 
\bibliography{ishikawa2022}

\end{document}